# NLP Sampling:
# Combining MCMC and NLP Methods for Diverse Constrained Sampling

Marc Toussaint* Cornelius V. Braun* Joaquim Ortiz-Haro†

July 4, 2024

**Abstract**

Generating diverse samples under hard constraints is a core challenge in many areas. With this work we aim to provide an integrative view and framework to combine methods from the fields of MCMC, constrained optimization, as well as robotics, and gain insights in their strengths from empirical evaluations. We propose NLP Sampling as a general problem formulation, propose a family of restarting two-phase methods as a framework to integrated methods from across the fields, and evaluate them on analytical and robotic manipulation planning problems. Complementary to this, we provide several conceptual discussions, e.g. on the role of Lagrange parameters, global sampling, and the idea of a Diffused NLP and a corresponding model-based denoising sampler.

## 1 Introduction

Sampling from a constrained set is a core problem in many areas: In robotics we may want to sample solutions subject to kinematic or physics constraints to tackle sequential manipulation planning. In Monte Carlo methods samples are generally used to estimate expectations. In generative models we generate samples to match a data distribution – analytical hard constraints have received less attention in this context so far.

In this work we consider how methods from across these field can be combined to provide strong constrained samplers. When starting this work we were in particularly interested in whether and how diffusion denoising approaches can be leveraged also for *model-based* hard-constrained sampling of diverse solutions to non-linear mathematical programs. However, our initial approach to this end failed, and Sec. 5 includes a conceptual discussion on this topic and our insights. Many other methodologies from across the fields can fruitfully be integrated in a joint framework, which is the focus of the technical contributions in this study.

[1]Learning & Intelligent Systems Lab, TU Berlin
[2]Machines in Motion Lab, New York University

To be more specific, we consider the problem of sampling from a distribution

$$p(x) \propto \exp(-f(x))\ \mathbb{I}_{g(x)\le 0}\ \mathbb{I}_{h(x)=0}, \tag{1}$$

where the indicator functions $\mathbb{I}$ impose elementwise non-linear constraints $g(x) \le 0$ and $h(x) = 0$. We assume the functions $f : \mathbb{R}^n \to \mathbb{R},\ g : \mathbb{R}^n \to \mathbb{R}^m,\ h : \mathbb{R}^n \to \mathbb{R}^{m'}$ are smoothly differentiable, and values and gradients can be queried point-wise. No a-priori data is available. To emphasize the analogy to non-linear mathematical programming (NLP), we we call this problem *NLP Sampling*.

Note that with only equality constraints, the problem becomes sampling on a non-linear (differentiable) constrained manifold, and with only linear inequalities, the problem becomes sampling from a polytope – our approaches will therefore incorporate ideas from existing specialized samplers for these cases. As is evident from these special cases, efficient NLP Sampling necessitates leveraging evaluations of the constraint functions $g(x), h(g)$ along with their gradients. This differentiates it from naively applying MCMC methods directly on $p(x)$, which has zero probability and score function in the infeasible space.

Constrained optimization can be highly efficient in generating solutions to problems with differentiable constraint functions in high-dimensional spaces, but are typically designed to only generate a single solution and are prone to local optima. Therefore, e.g. in the context of robotic path and manipulation planning, using constrained optimization as a component may render the overall solver brittle and lack probabilistic completeness [9, 20]. In contrast, classical MC methods as well as diffusion models and Stein-Variational Gradient Descent aim to generate a diversity of samples from a potentially multi-modal distribution.

In this work we first propose a family of *restarting two-phase NLP samplers* which is a framework to combine the best from across fields of constrained optimization, MCMC sampling, and robotics. We then discuss alternative approaches to the downhill phase (highlighting the strong relations, e.g. between Metropolis-Hasting and backtracking line search, or Riemannian Langevin and Gauss-Newton steps), as well as the interior sampling phase. We propose a novel interior sampling method, Non-linear Metropolis-Adjusted Hit-and-Run (NHR), which extends standard Hit-and-Run to non-uniform interior energies using Metropolis-Hasting as well as to non-linear inequalities. Additionally, we propose alternatives to informed restart seeding. As a basis for our evaluations we propose a novel metric to quantify diversity and mode coverage of samples, the Minimum Spanning Tree Score $\text{MSTS}_p$. Our evaluations highlight the weaknesses and strengths of the methods on analytical problems and robotics application, and conversely help us better understand the challenging non-linear and multi-modal structure of these problems.

Sec. 5 then provides several conceptual discussions, which are related to our evaluations but can also be read independently. This includes a discussion of *global* sampling (in analogy to global optimization), mode coverage and calibration, and how data-trained denoising ideas (which demonstrate global sampling) could be related to the problem of NLP Sampling. To make this concrete, we introduce the notion of a Diffused NLP, derive an approximation analogous to SQP methods, but mention our negative results with this first idea. The discussions include several pointers to future opportunities in NLP Sampling that further build on such ideas.

# 2 Related Work

**Robotic Path Planning.** Sampling-based methods have a long tradition in robotics. Prominently, sampling-based motion planning methods aim to find collision-free paths to a given goal under environmental constraints [13]. To this end these motion planners sample the high dimensional configuration space, using different strategies, until a path has been found [11]. While basic sample-based path planning does not consider differentiable feasibility constraints, Manifold RRT methods [18, 12] and other extensions account for such constraints. We will consider Manifold RRT as one candidate for what we will define as *interior* sampling, and thereby compare it to alternative MCMC methods, including Langevin and Hit-and-Run. However, as RRT methods are targeted to find continuous paths in configuration space, the sampling process expands continuously and cannot jump across infeasible regions to discover disconnected feasible regions. A core focus of our work is to also consider a phase-one sampling method to seed interior sampling within separated modes or disconnected regions of the feasible input space. Further, path planning methods are not directly comparable to MCMC methods, as they do not correctly sample from a given energy. By combining Manifold RRTs with Metropolis-Hasting we will bridge these two approaches.

**Robotic Task-and-Motion Planning.** Task-and-Motion Planning (TAMP; see [6] for a comprehensive survey) is a formulation of robotic manipulation planning, where discrete search over action sequences needs to be combined with continuous decisions about action parameters. The discrete action decisions impose feasibility constraints on the continuous decisions; and typical solvers need to generate continuous decisions subject to these constraints, e.g. using predefined samplers [7] or constrained optimization methods [19]. The present work was motivated by the core problem of generating diverse samples of feasible continuous decisions within TAMP, to ensure probabilistic completeness of the overall TAMP solver.

The idea of combining constrained optimization and samplers was previously discussed in [5, 15]. In the present work, we fully focus on the sub-problem of constrained sampling, combining the perspectives of MCMC and constrained optimization. The NLP samplers we propose here could be used as *streams* within PDDL-Stream [7], but also as a keyframe sampler within Logic-Geometric Programming [9, 20].

**MCMC, (Manifold-) Langevin, Polytope Sampling.** MCMC methods [1] are designed to sample from a given density $p(x)$; Langevin or Hamiltonian methods in particular can exploit a given score function (i.e. gradients are available). Some extensions of MCMC sampling methods to constrained manifolds exist [3]. However, some issues are insufficiently treated by such samplers: Manifold-MCMC methods typically assume the process is initialized feasible; the problem of finding a feasible initialization is neglected. Related to that, in cases involving disconnected manifolds (where MCMC steps across regions are highly unlikely), coverage and calibrated sampling across these modes is not neglected. Finally, these methods do not consider an inequality bounded feasible region, where we have interior space, in the terminology of constrained optimization. Our methods extend MCMC methods to address these issues.

Polytope sampling [4] addresses sampling from within a linearly bounded feasible region. This includes approaches that are interestingly different from typical Manifold-MCMC or Manifold-RRT methods, in particular Hit-and-Run (HR) [22]. We will propose a generalization of HR to account for non-linear inequalities (as well as equalities) and non-uniform interior density, and include it in our family of methods as an alternative to Manifold-RRT or -MCMC for interior sampling.

**Diffusion Denoising Models.** Probabilistic Diffusion Denoising Models (DDPMs) [10] use Langevin dynamics to generate samples. An important feature of the trained dynamics is that the generative process can cover multiple modes of the underlying density. This property makes it particularly interesting to what we want to achieve: a generative process to diversely sample from potentially disconnected constrained spaces. However, the power of Langevin dynamics to generate multi-modal samples in DDPMs has its origin in training on data that covers all the modes. So, therefore, the data contains the global information about the modes of the underlying distribution, and the trained denoising dynamics can exploit this knowledge to learn bifurcating dynamics that enables to sample from all modes. This view is in contrast to our setting, where we only have access to the underlying NLP (constraint and energy functions), being able to evaluate them point-wise, but without global knowledge or data about where modes or local optima might be. In this view, we aim to achieve the same efficiency and global coverage as data-trained DPPM sampling methods, but without availability of global data and based only on point-wise evaluations of the underlying problem. We will integrate Langevin dynamics (as well as Riemannian Langevin) as an option for NLP Sampling and empirically compare it to alternatives. Sec. 5 includes a more extensive discussion of the relations between NLP Sampling and denoising models.

**Algorithm 1** Restarting Two-Phase NLP Sampler

**Input:** Box limits $l, u \in \mathbb{R}^n$, differentiable functions $f, g, h$ can be point-wise queried

**Output:** Samples $D = \{\hat{x}_i\}_{i=1}^S$ with (approximately) $\hat{x}_i \sim \frac{1}{Z}\ e^{-f(x)}\ \mathbb{I}_{l \le x \le u}\ \mathbb{I}_{g(x) \le 0}\ \mathbb{I}_{h(x)=0}$

1: initialize $D = \emptyset$
2: **repeat**
3:     Seed $x \sim p_{\text{seed}}(D)$ (by default: $x \sim \mathcal{U}[l, u]$ uniform)
4:     **for** $k = 1, .., K_{\text{down}}$ **do**
5:         **if** $1^\top s(x) \le \epsilon$ **break**
6:         $x \leftarrow$ DOWNHILL$(x)$
7:     **end for**
8:     **if** $1^\top s(x) > \epsilon$ **restart** at line 3
9:     **for** $k = 1, .., K_{\text{burn}} + K_{\text{sam}}$ **do**
10:         **if** $k > K_{\text{burn}}$ and $1^\top s(x) \le \epsilon$ **then** $D \leftarrow D \cup \{x\}$
11:         $x \leftarrow$ INTERIORSAMPLE$(x)$
12:         **if** $1^\top s(x) > \epsilon$ **then** $x \leftarrow$ SLACKREDUCE$(x)$
13:     **end for**
14: **until** $|D| \ge S$

# 3 Restarting Two-Phase NLP Samplers

In the context of constrained optimization, "Phase I optimization" refers to the problem of finding a feasible point which can then, in Phase II, be used to seed an interior optimization method. We propose a family of sampling methods that equally adopts this basic two-phase approach. Specifically, a *Restarting Two-Phase NLP Sampler* iterates the following steps until enough samples are collected:

(i) Sample a new seed $x$, potentially conditional to previously found samples $D$.

(ii) A **slack downhill** method tries to find an (approximately) feasible point $x$ within $K_{\text{down}}$ steps.

(iii) If $x$ is feasible, an **interior sampling** method, potentially requiring $K_{\text{burn}}$ burn-in steps, tries to collect $K_{\text{sam}}$ samples $\sim e^{-f(x)}$ within the feasible space. If the method does not respect feasibility exactly, it is combined with a direct slack reduction (manifold projection) step.

Alg. 1 provides more explicit pseudocode, where $s(x)$ is a slack vector of constraint violation, defined in detail below. While straight-forward, we are not aware of sampling techniques to adopt this basic two-phase approach. We propose this family as a generic framework to integrate methods from across the fields of MCMC sampling, constrained optimization, and robotics. The following sections discuss alternatives for slack downhill and interior sampling within this framework.

## 3.1 Downhill, Noise, & Step Rejection on the Relaxed NLP

We first describe a class of methods that can be used for either slack downhill or interior sampling as combining a gradient-based downhill step with (optional) noise and an (optional) step rejection mechanism. This includes Langevin variants as well as classical constrained optimization techniques.

To introduce this class of methods we define the underlying energy as follows: Given constraint functions $g : \mathbb{R}^n \to \mathbb{R}^m,\ h : \mathbb{R}^n \to \mathbb{R}^{m'}$, we define $s(x) = ([g(x)]_+, |h(x)|) \in \mathbb{R}^{m+m'}$ as the *slack vector*, where the ReLU $[\cdot]_+$ and absolute value $|\cdot|$ are applied element-wise. This is the stacked vector of all inequality and equality violations. The total violation is given as $1^\top s(x)$, which is also the $L_1$-slack penalty. However, in this work we focus on the $L_2$ slack penalty $s(x)^\top s(x)$ to exploit its Gauss-Newton steps. Given an interior energy $f : \mathbb{R}^n \to \mathbb{R}$, we define the *relaxed NLP* as

$$F_{\gamma\mu}(x) = \gamma f(x) + \mu s(x)^\top s(x)\ . \tag{2}$$

While introducing two scalings $\gamma$ and $\mu$ is an unusual convention, we want to explicitly include $\gamma = 0$ to describe pure slack downhill (Phase I) as $F_{01}$, and interior sampling as $F_{1\mu}$ for large $\mu \gg 1$. This relaxed NLP also implies the *relaxed sampling problem* $p_{\gamma\mu}(x) \propto \exp\{-F_{\gamma\mu}(x)\}$, so that we can consider various MCMC methods for both phases, slack downhill ($p_{01}$) and interior MCMC ($p_{1\mu}$).

For a given an energy $F(x)$, Appendix A recaps the definitions of Langevin steps, Riemannian Langevin, Metropolis-Hasting (MH), Metropolis-Adjusted Langevin (MALA), Riemannian MALA, Gauss-Newton steps, and the Armijo rule of backtracking line search. We decided to move this textbook knowledge to the appendix, but want to highlight the strong relations between these methods here, some of which are obvious and well-known, some less. For instance, in the appendix we observe: (1) Discrete time Langevin dynamics (13) is the same as combining plain gradient steps $-\alpha\delta$ with isotropic noise $\sigma z$, $z \sim \mathcal{N}(0, \mathbf{I}_n)$, where the noise scaling $\sigma = \sqrt{2\alpha}$ is tied to the step size $\alpha \equiv \tau$. (2) Discrete time Riemannian Langevin dynamics (14) [8] is the same as combining Newton downhill steps $-\alpha H^{-1}\delta$ with covariant noise $\sigma\sqrt{H^{-1}}z$, $z \sim \mathcal{N}(0, \mathbf{I}_n)$, where the noise scaling $\sigma = \sqrt{2\alpha}$ is still tied to the step size $\alpha \equiv \tau$.

While the first observation is often mentioned, the direct relation between Riemannian Langevin and Newton methods is, in the original paper, not mentioned. Both observations render discrete time Langevin dynamics to be a special case of classical downhill with adding noise, where the noise scaling is tied in a very particular way to the choice of step size. Tying the noise scale to the step size is at the core of (continuous time) Langevin's property to actually generate samples of the underlying distribution – which is relevant for interior sampling, but less for downhill.

Discrete time Langevin needs to be combined with Metropolis-Hasting (MH, see Eq. 17) to correctly sample from an underlying distribution $p(x)$ – this is termed Metropolis-Adjusted Langevin (MALA). MH essentially rejects non-decreasing steps with some probability. It turns out that there is a strong relation to the Armijo rule used in classical backtracking line search. In the appendix we observe: (3) The acceptance rate (21) of a gradient downhill step $x' = x - \alpha\delta$ under Metropolis-Hasting differs from the Armijo rule with line search parameter $\rho = \frac{2\alpha}{\sigma^2}$ only in that it uses the more symmetrical gradient estimate $\bar\delta$ instead of $\delta$, and accepts with some probability $< 1$ also steps with $F(x') > F(x) - \frac{2\alpha^2}{\sigma^2}\delta^2$.

In view of these analogies we think of all mentioned methods as an instance of the family of methods that combines three steps:

(i) We can use plain gradient or the Newton direction (2 options) to make a downhill step scaled by $\alpha$;

(ii) we can add isotropic or covariant or no noise (3 options) scaled by $\sigma$; where $\sigma$ might be tied to $\alpha$ for Langevin variants;

(iii) and we can reject the overall step based on the Armijo rule or Metropolis-Hasting or always accept (3 options).

Any of these methods can be applied on $F_{01}$ for slack downhill, or $F_{1\mu}$ for interior sampling. Our evaluations will systematically compare them in both roles. To precise we add that, in our applications, we always approximate Newton steps for $s(x)^\top s(x)$ as Gauss-Newton, i.e.:

$$(\nabla^2 F_{\gamma\mu}(x) + \lambda\mathbf{I})^{-1}\nabla F_{\gamma\mu}(x) = (\gamma\nabla^2 f(x) + 2\mu J_s^\top J_s + \lambda\mathbf{I})^{-1}(\gamma\nabla f(x) + 2\mu J_s^\top s) \ , \tag{3}$$

where all RHS terms ($s$ and $J_s$) are evaluated at $x$ and $\lambda$ is a Levenberg-Marquardt damping (or Tikhonov regularization) parameter. Further, all stepping methods are subject to box bounds $l, u \in \mathbb{R}^n$ and a potential maximum step length $\delta_{\max}$.

It might surprise that Lagrange methods play no role in what we propose for NLP Sampling. Sec. 5.1 discusses this in more depth.

### 3.2 Interior Sampling with Explicit Constrained Handling

The previous section described methods that operate on a differentiable energy $F(x)$ – akin to unconstrained optimization. In contrast, in this section we consider sampling methods that deal with constraints

**Algorithm 2** Standard Hit-and-Run [22]

**Input:** Inequalities $g(x) = Gx + \hat{g} \le 0$, max step $\delta_{\max}$, initial (near-)interior point $x$

1: **repeat**<br>
2: sample random direction $d = z/|z|$ with $z \in \mathcal{N}(0, 1)$<br>
3: initialize $[\beta_{\text{lo}}, \beta_{\text{up}}] = [-\delta_{\max}, \delta_{\max}]$<br>
4: let $\bar{g} \leftarrow Gx + \hat{g}$, $a \leftarrow Gd$ (so that $g(x + \beta d) = \bar{g} + \beta a$)<br>
5: clip: $\forall_{i:a_i<0} : \beta_{\text{lo}} \leftarrow \max\{b_{\text{lo}}, -\bar{g}_i/a_i\}$ , $\forall_{i:a_i>0} :$ $\beta_{\text{up}} \leftarrow \min\{b_{\text{up}}, -\bar{g}_i/a_i\}$<br>
6: step: $x \leftarrow x + \beta d$ with $\beta \sim [\beta_{\text{lo}}, \beta_{\text{up}}]$ uniform<br>
7: **until** forever

**Algorithm 3** Non-linear Metropolis-Adjusted Hit-and-Run (NHR)

**Input:** Non-linear inequalities $g(x)$, energy $f(x)$, max step $\delta_{\max}$, box bounds $l, u \in \mathbb{R}^n$, initial (near-)interior point $x \in \mathbb{R}^n$

1: **repeat**<br>
2: sample random direction $d = z/|z|$ with $z \in \mathcal{N}(0, 1)$<br>
3: initialize $[\beta_{\text{lo}}, \beta_{\text{up}}] = [-\delta_{\max}, \delta_{\max}]$<br>
4: clip $[\beta_{\text{lo}}, \beta_{\text{up}}]$ with lower ($\bar{g} = l - x, a = -d$) and upper ($\bar{g} = x - u, a = d$) bounds<br>
5: clip $[\beta_{\text{lo}}, \beta_{\text{up}}]$ with $\bar{g} = g(x), a = J_g(x)d$ *// is optional!*<br>
6: **repeat**<br>
7: **if** $\beta_{\text{lo}} > \beta_{\text{up}}$ **then** break *// failure, no step*<br>
8: $y \leftarrow x + \beta d$ with $\beta \sim [\beta_{\text{lo}}, \beta_{\text{up}}]$ uniform<br>
9: linearize $g^y = g(y), G^y = J_g(y)$<br>
10: **if** $g^y \le 0$ **then**<br>
11: with probability $\min\{1, \frac{\exp\{-f(y)\}}{\exp\{-f(x)\}}\}$ accept $x \leftarrow y$<br>
12: break<br>
13: **end if**<br>
14: only for $i$ with $g_i^y \ge 0$: clip $[\beta_{\text{lo}}, \beta_{\text{up}}]$ with $\bar{g} = g^y + G^y(x - y)$ and $a = G^y d$<br>
15: **until** forever<br>
16: **until** forever

more explicitly. We first propose an extension of Hit-and-Run – an interesting approach to deal with inequalities – to become applicable to our general NLP Sampling problem, and then discuss non-linear manifold RRT methods – an interesting approach to deal with equalities.

### 3.2.1 Non-linear Metropolis-Adjusted Hit-and-Run (NHR)

A set of linear inequalities $Gx + \bar{g} \le 0$ defines a polytope. Several approaches to sampling uniformly from a polytope are well-established [4] and can easily be adjusted using MH to also account for an interior target density $\propto e^{-f(x)}$. A standard family of approaches are Ball and Dikin Walks, which use step proposals sampled uniformly from a ball or ellipsoidal and reject them when outside the polytope [4]. Dikin Walks are very similar to Riemannian Langevin on $F_{1\mu}$ for $\mu \to \infty$.

However, in this section we are particularly interested in another approach as it is quite different to typical MCMC walks with local proposals: Hit-and-Run (HR) [22]. HR samples a random direction in $\mathbb{R}^n$ (uniformly), computes where this line intersects the inequalities, and then samples uniformly from the interior line segment. Alg. 2 defines standard HR.

We extend HR to become applicable to non-linear inequalities, as well as MH-adjust it to non-uniform interior $p(x) \propto e^{-f(x)}$. Alg. 3 presents the novel method. All the clip operations use line 5 of the original Hit-and-Run for some $\bar{g}$ and $a$. The inner loop (line 6) samples a candidate $y$ on the line, evaluates the inequalities $g(y)$, and if they are violated clips the interval $[\beta_{\text{lo}}, \beta_{\text{up}}]$ with only the violated inequalities using the linearization at $y$. Note that the initial clipping (line 5) with the linearization at $x$ is not necessary, which leads to more exploration.

**Algorithm 4** Manifold-RRT (mRRT), following [18]

**Input:** box limits $l, u \in \mathbb{R}^n$, differentiable functions $f, g, h$ can be point-wise queried

$N$ is an approximate Nearest Neighbor data structure, where we store tuples $(x, P_x)$ with tangent projection matrix $P_x$
$x$ is seed near manifold (e.g., from Phase I)
**for** $f = 1, .., K_{\text{sam}}$ **do**
  **if** $1^\top s(x) \le \epsilon$ **then** $D \leftarrow D \cup \{x\}$
  $P_x \leftarrow \mathbf{I} - J_h^\top (J_h J_h^\top + \epsilon)^{-1} J_h$
  $N \leftarrow N \cup \{(x, P_x)\}$
  $\hat{x} \sim \text{uniform}[l, u]$
  $(x, P_x) \leftarrow N.\text{nearest}(\hat{x})$
  $\delta = P_x(\hat{x} - x)$, $\delta \leftarrow \frac{\alpha}{|\delta|}\delta$
  $x \leftarrow x + \delta$
  $x \leftarrow \textsc{SlackReduce}(x)$ (also to deal with inequalities)
**end for**

We can apply NHR on problems with equality constraints with two modifications: First, the random direction $d$ is chosen tangential to the equalities $h$, as given by the projection $(\mathbf{I} - P_h)$ with $P_h = J_h^\top (J_h J_h^\top)^{-1} J_h$. Second, equalities are turned into $\epsilon$-margin inequalities. To get samples exactly on the manifold we need to post-process samples with slack steps.

### 3.2.2 Non-linear Manifold Rapidly-Exploring Random Tree (mRRT)

RRTs generate samples by gradually growing from an initial sample (start configuration) towards uniformly covering the sample's connected component [13]. The incremental nature (within a connected component) is analogous to MCMC walks, but the fact that it generates a tree instead of a chain, i.e., grows "at all leafs" makes it explore fast and an interesting candidate for sampling in general, regardless that it was developed to return connected paths.

RRTs have been generalized to grow on non-linear manifolds [18]. All these approaches require an initial sample on the manifold to grow from, which our Phase I step can provide. For completeness, we include our specific implementation of manifold RRT in Alg. 4. In this algorithm, growth steps are projects to be tangential to the equality constraints. A following SlackReduce step has to correct for non-linearities in the constraints, but also corrects for inequality violation by projecting to the nearest feasible point.

## 3.3 Conditional Restart Seeding

We discussed various alternatives for the slack downhill and interior sampling phases. (For slack reduce we will always use a Gauss-Newton slack step with full step length $\alpha = 1$.) The seeding of restarts conditional to previously found samples $D$ is yet open.

We consider three options, where the last is, to our knowledge, novel:

(i) Sample $x \sim \mathcal{U}[l, u]$ independent of previous samples, which makes all episodes i.i.d.

(ii) Sample a candidates $\{x_1, .., x_C\}$ uniformly and select $\text{argmax}_i \min_{y \in D} |y - x_i|$, which maximizes the nearest distance to the data $D$.

(iii) Sample a candidates $\{x_1, .., x_C\}$ uniformly, compute the Gauss-Newton slack step $\delta_i$ for each, and select $\text{argmin}_i \max_{y \in D} \frac{(y - x_i)^\top \delta_i}{|y - x_i|\,|\delta_i|}$, which minimizes the alignment of the slack direction with any data direction.

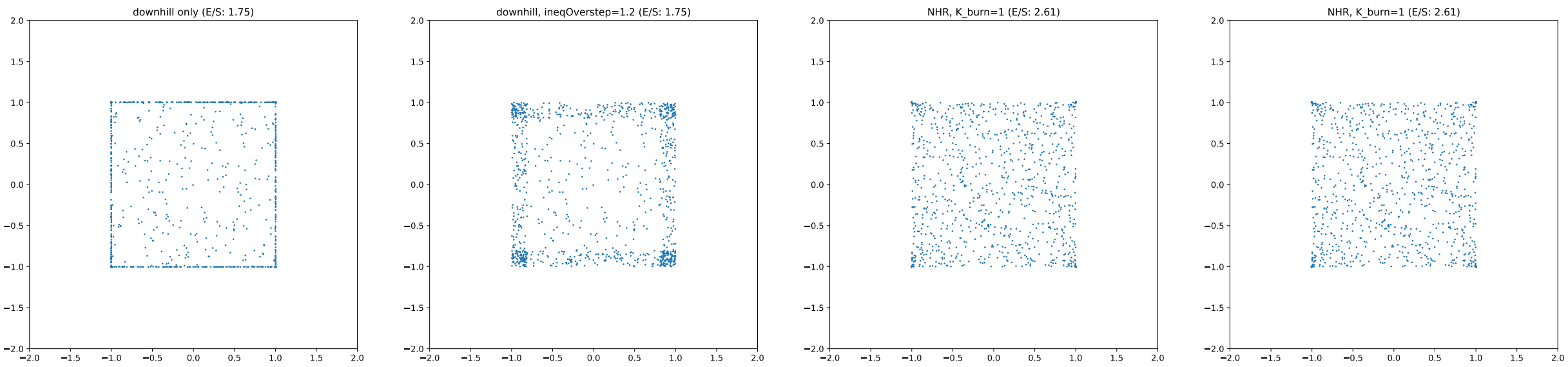

(a) Pure GN downhill without interior sampling

(b) Downhill with over-stepping GN steps ($\alpha = 1.2$)

(c) Pure downhill with single NHR interior sampling burn-in step ($K_{\text{burn}} = 1$)

(d) Same with $K_{\text{burn}} = 2$

Figure 1: Samples generated by different downhill and interior methods.

# 4 Empirical Evaluations

The full sources to reproduce all evaluations are available here.[1]

## 4.1 Evaluation Metrics and Minimum Spanning Tree Score (MSTS)

We generally found it hard to find a useful metric that reflects a degree of diversity in the sampled dataset $D$. For a low-dimensional toy problem we can use the Earth Mover Distance[2] between the samples and ground truth samples (e.g. generated using naive rejection sampling) to check the "correctness" of sampling. However, in the problems of interest to us we care less about correctness after asymptotic mixing, but rather the rate of diversity and mode coverage we get with each and early samples.

We first considered metrics that analyze the neighborhood of samples, e.g. the spectrum of distances to the ten nearest neighbors. However, such metrics are fully ignorant on whether different modes are populated. Nearest neighbor-based metrics can capture uniformity or how locally equidistant samples are drawn, but not coverage of modes.

We found the size of the minimum spanning tree of all points in $D$ an interesting quantity that does capture also distances between covered modes, as at least one edge needs to connected distant covered modes. Specifically, we define the Minimum Spanning Tree Score $\text{MSTS}_p(D)$ as the total cost of the tree when using edge costs $|x - x'|^p$.

When edge costs are plain Euclidean distances, $p = 1$, it is guaranteed that $\text{MSTS}_1(D)$ strictly increases with $n = |D|$ when adding points. Interestingly, the rate of increase seems to give an indication of the dimensionality of the sampled space: E.g., when grid-sampling $n$ points from a box of dimension $d$ we expect an $\text{MSTS}_1(D)$ of roughly $n^{1-1/d}$ (the number of lines of length 1 along one dimension of the grid). Our experiments suggest that a similar behavior is true for uniformly sampled points.[3]

With edge costs $|x - x'|^p$ for $p > 1$, the $\text{MSTS}_p(D)$ does not monotonously increase with $|D|$ when adding points. Instead, with increasing $|D|$ modes become more and more densely populated, with neighbor distances going to zero and contributing less to MSTS. However, distances between *separated* modes stay greater than zero and add up. Therefore, $\text{MSTS}_p(D)$ will (for large $|D|$) sum over mode separations and become a metric of separateness of modes.

Using restarts and interior sampling, our methods generate the sample dataset $D_n = \{x_i\}_{i=1}^n$ sequentially. Given the sequence $x_i$, $i = 0, .., n$, we compute $\text{MSTS}_p(D_n)$ for each data subset $D_n$ so that we can plot

[1] https://github.com/MarcToussaint/24-NLP-Sampling

[2] We use OpenCV's implementation.

[3] We would be interested in theory on the rate of $\text{MSTS}_p$ with $n$ depending on the dimensionality or topological properties of the feasible space, or when the dimensionality varies as in simplicial complexes.

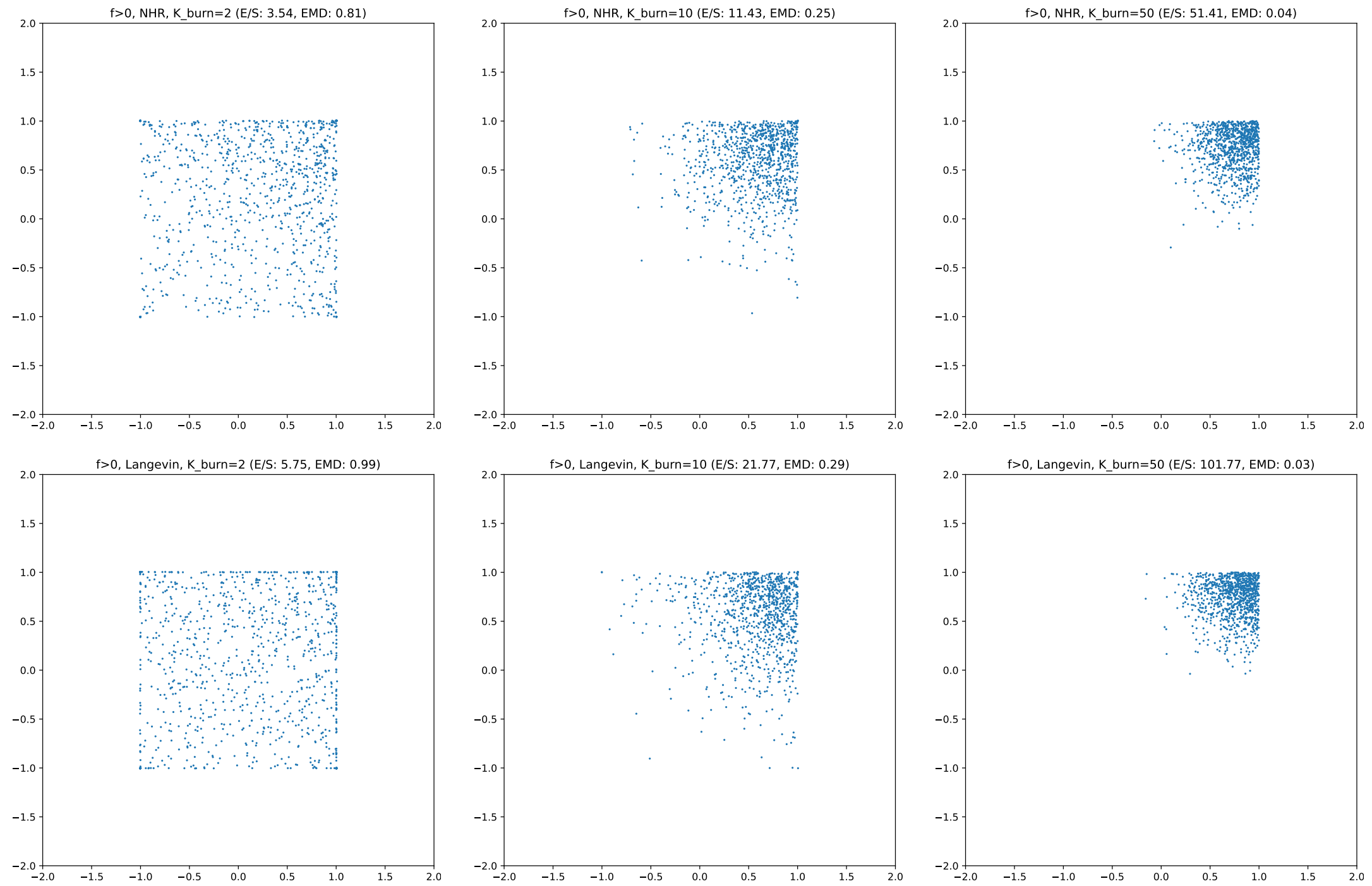

Figure 2: NHR and Langevin for sampling from a box with interior energy $f = (x - (1, 1))^2$, for varying $K_{\text{burn}}$. Captions report on Evaluations/Sample and Earth Mover Distance to ground truth. Plain MCMC is visually also very similar, but converges somewhat slower. (Evaluations per sample are roughly proportional to $K_{\text{burn}}$ as only a single sample $K_{\text{sam}} = 1$ is taken after burn-in.)

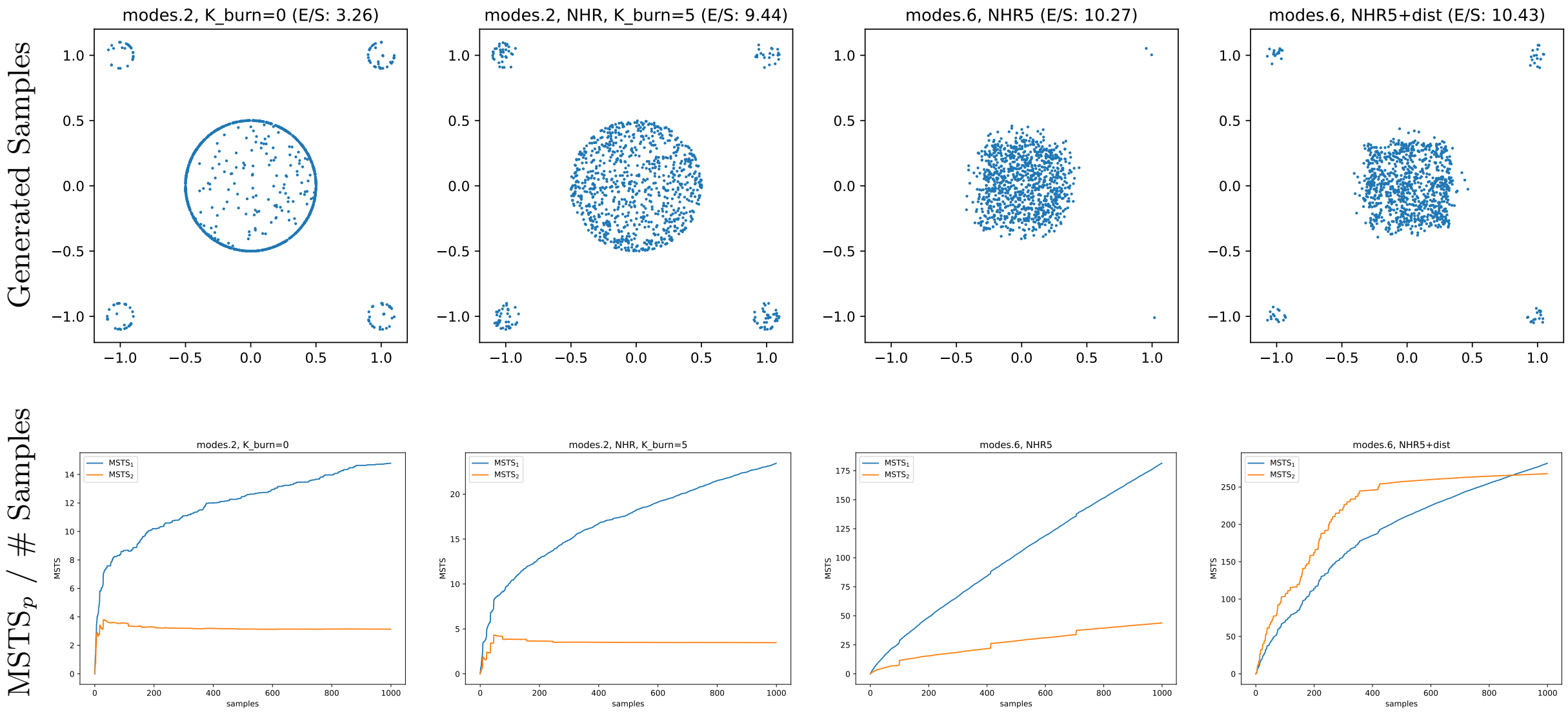

(a) Pure GN downhill on the 2D modes problem

(b) NHR ($K_{\text{burn}} = 5$) on the 2D modes problem

(c) Pure GN downhill on 6D modes problem

(d) NHR ($K_{\text{burn}} = 5$) on the 6D modes problem

Figure 3: Performances on the modes problem. The top row shows generated samples in 2D (a-b) and 6D (c-d). The bottom row shows $\text{MSTS}_p$ with growing #samples.

the score as a function of $n$ and the number of evaluations needed to generate $n$ samples.

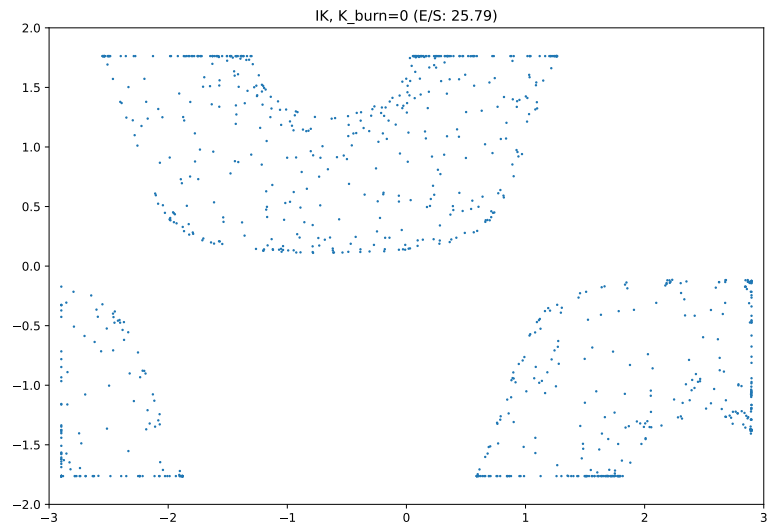

(a) Pure GN downhill

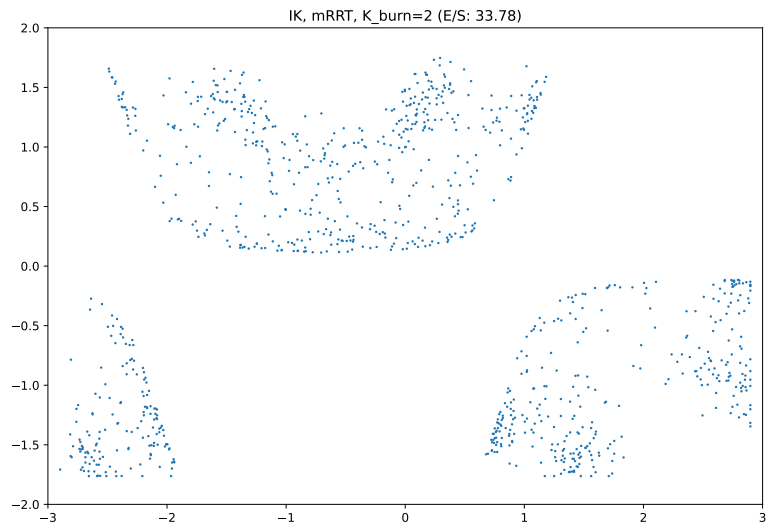

(b) mRRT with $K_{\text{burn}} = 2$ burn-in steps

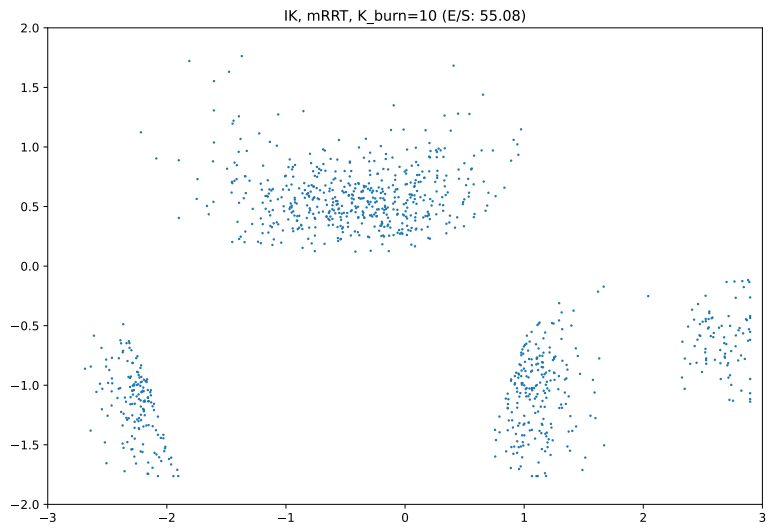

(c) mRRT with $K_{\text{burn}} = 10$

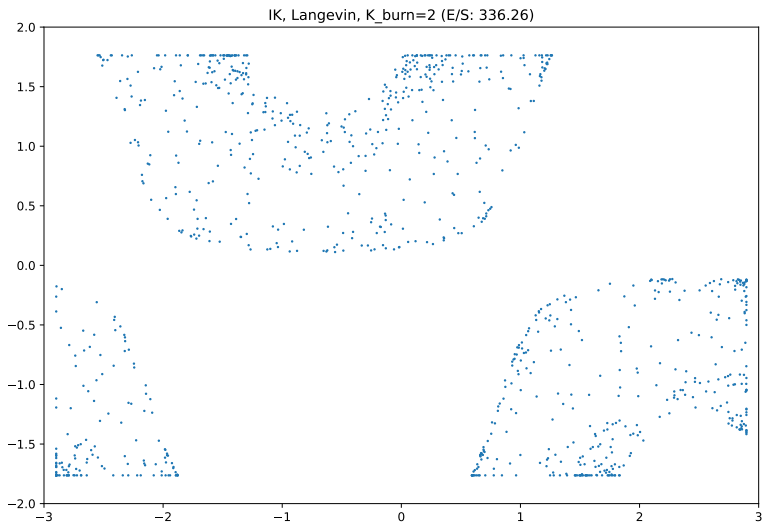

(d) Langevin with $K_{\text{burn}} = 2$

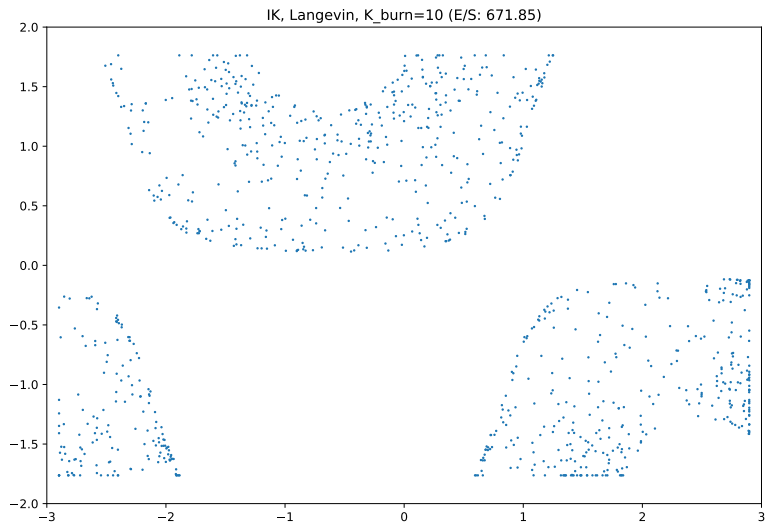

(e) Langevin with $K_{\text{burn}} = 10$

Figure 4: Samples generated by different downhill and interior methods for the Inverse Kinematics problem ($x \in \mathbb{R}^7$, only the first two dimensions displayed). Note the high evaluations per sample for Langevin methods in this highly non-linear setting.

## 4.2 Illustrations

We start with illustrations of the methods' behavior, which is best seen in scatter plots of samples. We first consider the uniform box problem in 2D, with 4 inequalities $-(x+1) \le 0,\ x-1 \le 0$ (elem-wise), flat energy $f(x) = 0$, and bounds $l = (-2,-2), u = (2,2)$ (which are known and used for seed sampling). Fig. 1 shows how pure downhill generates too many samples at the boundary; and that overstepping Gauss-Newton (GN) downhill steps can help to jump into the interior rather than staying at the border. The third and fourth plots illustrate how efficiently NHR mixes within the box with just $K_{\text{burn}} = 1$ or $K_{\text{burn}} = 2$ burn-in steps.

To check correctness in handling non-flat energies we added $f = 4(x-1)^2$ to the 2D box problem, which is a Gaussian around $\mathbf{1}$, but clipped by the inequalities. Fig. 2 shows the results when using NHR, plain MCMC, or Langevin as interior sampling method, and $K_{\text{burn}} \in \{2, 10, 50\}$ burn-in steps. The Earth Mover Distance confirms convergence to the true distribution for all three methods.

To illustrate mode exploration and the $\text{MSTS}_p$ score, we define the *modes problem* in $n$ dimensions as having $1 + 2^n$ centers $c_{0,..,2^n} \in \mathbb{R}^n$ and radii $r_{0,..,2^n} \in \mathbb{R}$. Together they define a highly non-linear single inequality function

$$g(x) = \min_{i \in \{0,..,2^n\}} (x - c_i)^2 / r_i^2 - 1 \ , \tag{4}$$

which describes a feasible space $\{x : g(x) \le 0\}$ which is composed of balls around each $c_i$ with radius $r_i$. We fixed $c_0 = 0$ to be at the origin, and each $c_i, i = 1, .., 2^n$ to be at the corner $(\pm 1, \pm 1, .., \pm 1)$ with signs the binary code of $i$. We further fixed the center mode to have radius $r_0 = 0.5$, and all other modes $r_i = 0.1$. The range bounds are $l = -1.2, u = 1.2$.

The two leftmost columns in Fig. 3 show results on the 2D modes problem, for pure downhill and downhill as well as $K_{\text{burn}} = 5$ NHR steps. The $\text{MSTS}_p$ plots show that $\text{MSTS}_1$ continuously increases (more so when using NHR), while $\text{MSTS}_2$ converges to a constant, indicating total mode separateness. The two columns on the right show results for the 6D modes problem, where we have $1 + 2^6 = 65$ modes. The same method (using NHR with $K_{\text{burn}} = 5$) here struggles to discover modes; the occasional steps in the $\text{MSTS}_2$ score clearly indicate newly found modes. However, when using $\text{dist}_{100}$ restart seeding modes are robustly discovered and also the $\text{MSTS}_1$ scores increase more rapidly.

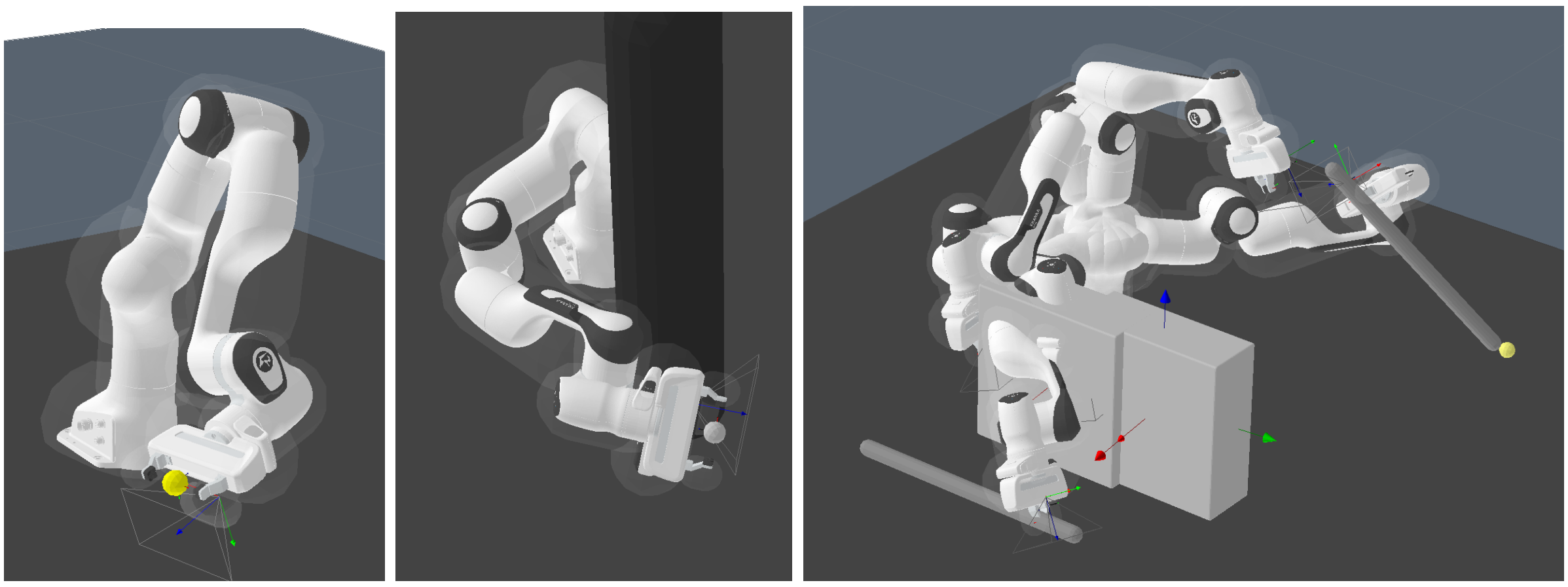

Figure 5: a) IK problem: reaching for the yellow target; b) similar, with a large cylindrical obstacle (implying non-linear collision inequalities and disconnected modes), c) a complex sequential manipulation planning problem over 4 consecutive configurations: start-of-box-push, end-of-box-push, reach-stick, and touch-target-with-stick – overlaid in this solution illustration. See text for details.

We also illustrate results for a 7D robotics inverse kinematics problem (details below), where a non-linear equality constrains the endeffector position, and inequalities model joint limits and collisions. Fig. 4 provides scatter plots of the first two dimensions (robot arm joints) only. The first plot shows again concentration of pure downhill at the border of the feasible space (in 7D, samples may only *appear* interior in the 2D projection); the 2 subsequent plots illustrate mixing by mRRT interior sampling (quantitatively evaluated below). The two bottom plots show Langevin interior sampling; note the high Evaluations/Sample in this highly non-linear equality constrained problem.

### 4.3 Benchmarks Problems

We evaluate the methods on 5 analytic problems, and 3 robotic problems. The 5 analytic problems are the uniform box in 2D and 6D, the "modes problem" in 2D and 6D, and random linear programs (LPs) in 2D. For the random LPs, we have a linear inequality $g(x) = Gx - 0.2$ with $G \in \mathbb{R}^{5n \times n}$, and we sample $G$ normal Gaussian.

Concerning the robotics problems, we considered a standard Inverse Kinematics (IK) problem with a 7-dof Franka arm, there the goal is to reach reference point (yellow in Fig. 5a) with the endeffector; then a similar reaching problem but with a massive cylindrical obstacle (defining distance inequalities to all robot collision shapes) which can be circumvented to the left or right. For both of these problems, the feasible space is "large" in the sense that it contains many (also unintuitive) feasible configurations, but is dominated by the highly non-linear equality constraint and the many inequalities for the cylinder problem. The third robotics problem is a sequential manipulation planning problem where we jointly solve for 4 consecutive configurations, which are the start-of-box-push, the end-of-box-push, the reach-stick, and the touch-target-with-stick configurations. The story is that the target can only be reached with a stick, which is initially behind a box that first needs to be pushed away. Fig. 5c) illustrates a solution by overlaying all 4 configurations – this looks cluttered but is actually a very clean and nice solution, where the box is properly pushed aside to make space for the stick reach. The problem is 49-dimensional (larger than $4 \cdot 7$, as many additional parameters such as box placement, relative push pose, and stick grasp are co-optimized decision variables), and highly non-linear in all equalities and inequalities. The box pushing is modeled as stable push[4]. The feasible space is again "large", e.g., where the box is placed is fully open, but constrained in a complex manner. Pure downhill gets often stuck in infeasible local slack minima and without diverse solutions, downstream robotic solvers (path planners or trajectory optimization, [9]) struggle to find globally appealing solutions.

[4]The `robotic` python package tutorial includes this push model.

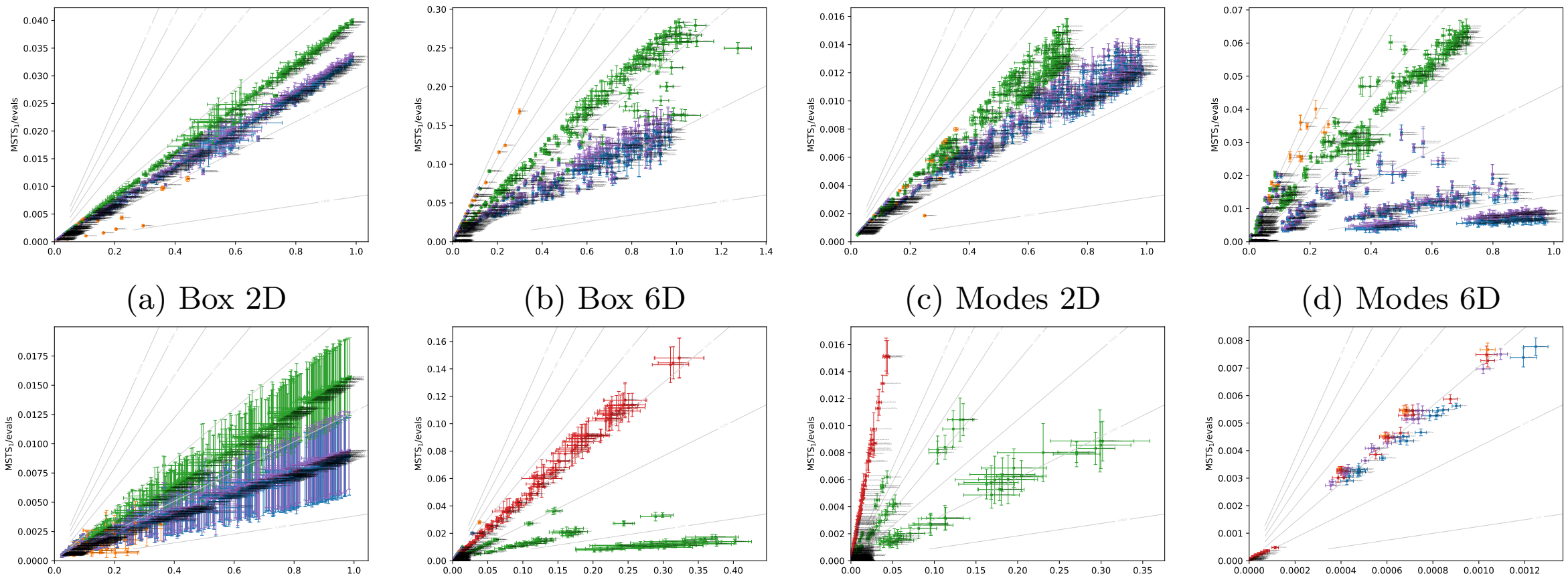

(a) Box 2D (b) Box 6D (c) Modes 2D (d) Modes 6D

(e) Linear Programs (f) Inverse Kinematics (g) Cylinder Obstacle (h) Push

Figure 6: Performance plots for all method combinations on the 8 problems. The two axis ($MSTS_1$-per-evals and samples-per-evals) show better methods towards the top right; angular isolines ($MSTS_1$-per-sample) show better method appear at higher "polar angle". The error bars show standard deviation over the 10 runs (linear programs are themselves randomized). The color of performance crosses indicates the interior sampling method (orange: none, green: NHR, blue: MCMC, red: manifoldRRT, violet: Langevin).

## 4.4 Results

Fig. 6 displays results from running over 172 800 experiments, which cover 8 problems, 1 584 tested method combinations, and 10 runs for each of these combinations. The methods combine a downhill method (`GN, grad`), downhill noise method (`none, iso, cov`), downhill reject method (`none, Wolfe, MH`), interior sampling method (`NHR, mRRT, MCMC, Langevin`), interior sampling burn in steps (`0, 5, 20`), interior samples taken (`1, 5, 20`), and the restart seed method (`uni, nov, dist`). Some nonsensical combinations are excluded, such as Metropolis-Hasting without noise, or various interior sampling methods while only taking a single sample. We put a limit $K_{\text{down}} = 50$ on downhill steps throughout, except for the push problem where we choose $K_{\text{down}} = 200$.

Fig. 6 uses two performance metrics ($MSTS_1$-per-evals and samples-per-evals) to position all methods in a 2D performance space, where the performance metrics were computed after a full run (after either 1000 samples are collected or 100 000 evaluations were exceeded). Here 'samples' means the number of returned (feasible) samples, and samples-per-evals is a number in $[0, 1]$. These performance plots show better methods towards the top right. The plots show that groups of methods (with same interior sampling method) seem to appear on lines in this 2D performance plot. Note that the slope of these lines relates to $MSTS_1$-per-sample, which relates to the diversity of returned samples. We therefore added isolines for this third performance indicator; where better method appear at higher "polar angle". The tiny black flags indicate the method details for each performance cross.

We see interesting trends in these full results. However, as plots are rather cluttered, we provide Fig. 7 to aggregate these results into a series of head-to-head method comparisons. This will allows us to sub-select promising methods to inspect and discuss their performance plots in more detail below. We summarize our findings in Fig. 6 and Fig. 7 as follows:

- A clear finding is that burn-in steps for interior sampling do not pay off in these metrics – which makes sense as our metrics are relative to # evaluations needed to generate the samples. (But certain application scenarios might justify burn-in steps to warrant sample quality.)

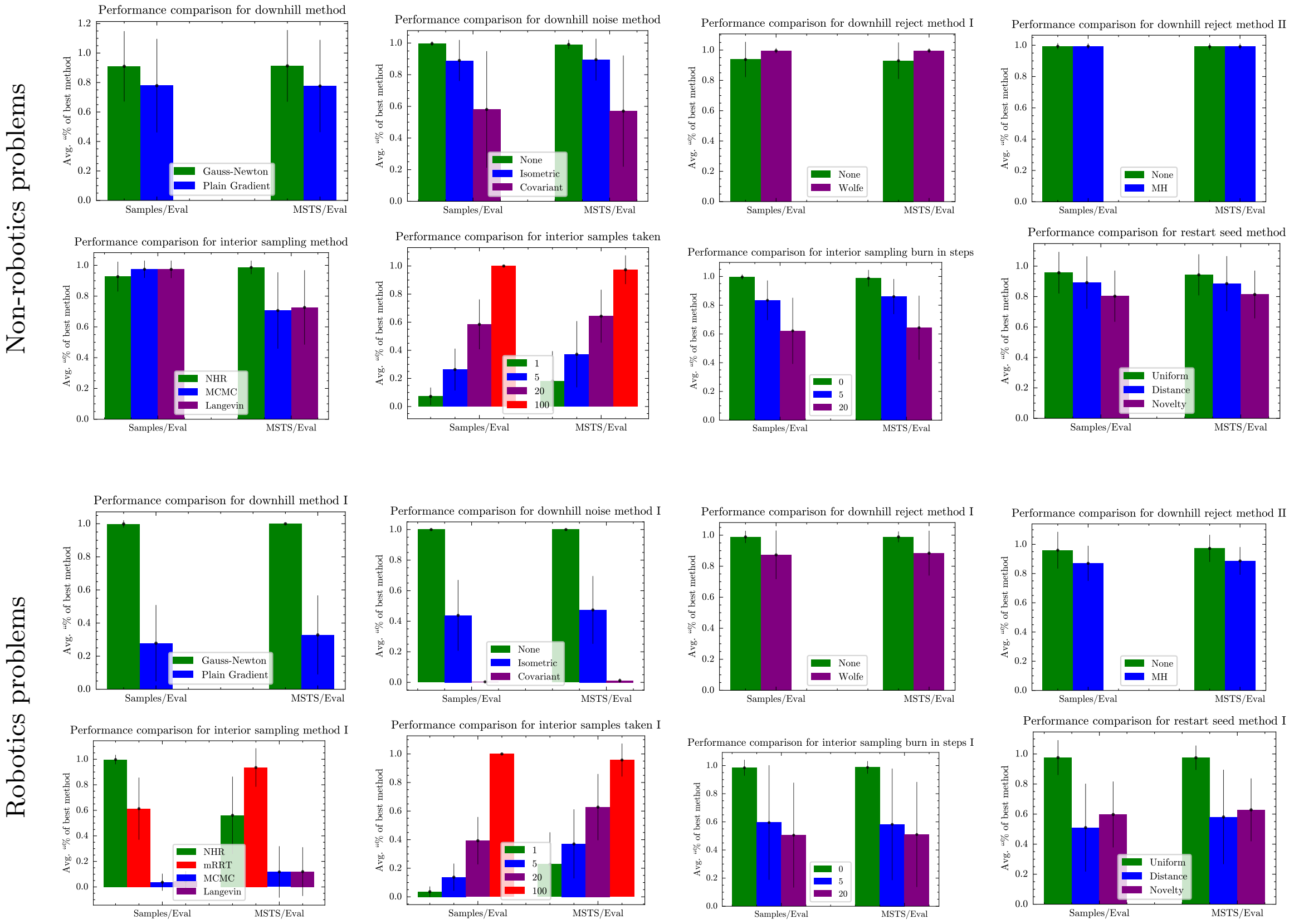

Figure 7: Method comparison across all problems. Each plot shows the average performance of each method relative to the per-problem-best performing method (i.e. 1.0 if the very method is the best for a problem / method combination). This means that only exact matching combinations are compared, that is for instance GN with covariant noise (and a specific restart method etc.) vs. gradient with covariant noise and same other set-up. Due to their different properties (e.g. equality constraints or not) we split the aggregation into robotics (bottom) and non-robotics problems (top).

- We also observe that Gauss-Newton downhill steps generally outperform plain gradient steps in Phase I. This performance difference is exacerbated on the robotics problems.
- Using downhill reject methods (`Wolfe, MH`) are regularly among the best methods. However, the difference is small. We explain this due to the important role of restarting: Single-run non-linear optimization and MCMC methods require Wolfe and MH to guarantee monotone convergence or probabilistic correctness. However, our samplers, if they happen to be stuck, will be restarted and the results indicate that aggressive downhill eventually pays off in terms of diversity and samples per evaluations.
- Injecting noise during downhill does sometimes lead to good methods, but no noise is generally very competitive. In particular, we observe that injecting covariant noise tends to perform worse than isometric noise.

Based on the above observations from Fig. 7, we reduce clutter in the plots by excluding method combinations accordingly. Fig. 8 provides the performance plots of the remaining methods for six problems. Our findings are:

- Focusing on the color coding first, we find that NHR (green) interior sampling performs particularly well on linear programs (with high variance due to the randomized problems), including the box problem, but also the modes problem. In turn, on the robotics problems (dominated by hard equality constraints instead), mRRT (red) shows its strength. Esp. on these problems plain MCMC (blue) and Langevin (violet) are not efficient, as they are not manifold constrained.

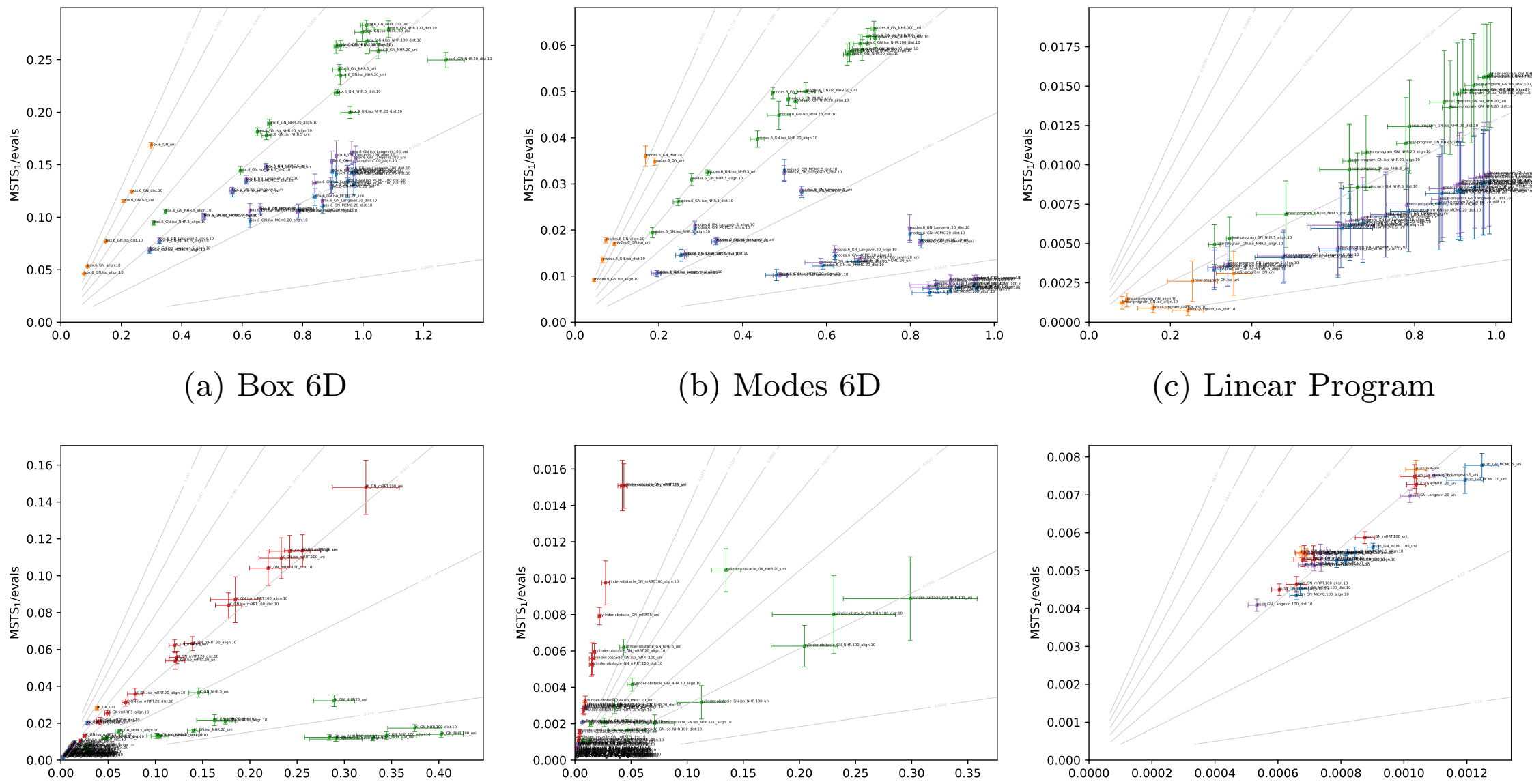

(a) Box 6D (b) Modes 6D (c) Linear Program

(d) Inverse Kinematics (e) Cylinder Obstacle (f) Push

Figure 8: Selected methods are positioned in 2D performance plots, with better methods appearing top right with high diversity($\mathrm{MSTS}_1$)-per-eval and high samples-per-eval. Note that the polar angle of the 2D placement relates to diversity-per-sample, which is also of interest and better methods appear top left. See Sec. 4.4 for details on the performance plots. Performances for the 8 problems are given. The color code indicates the used interior sampling method (orange: none, green: NHR, blue: MCMC, red: manifoldRRT, violet: Langevin), the labels provide details of parameter settings. See Appendix Fig. 6 for plots with all methods.

- Looking into the restart seeding (annotated with uniform, distance-based, and novelty-based), we find that esp. for the modes problem, diverse seeding can increase $\mathrm{MSTS}_1$-per-eval. However, especially for the robotics problems, diverse seeding seems to be clearly harmful! This came to a surprise, and drastically changed our conception of the structure of these problems: Apparently, seeding away from previously found feasible samples is harmful. This suggests that the feasible space is actually quite local in regard to the high-dimensional embedding space (49-dimensional, in the push example), but at the same time highly multi-modal (esp. for the cylinder and push problem) and non-linear.

  Concerning the alignment-minimizing seeding, the method leads to good diversity-per-sample (the polar coordinate) in the box, modes, and LP problems. But when evaluating the performance *per evaluations* the method is far behind plain uniform and distance-based reseeding: Clearly the evaluations needed to evaluate alignment of seed candidates are too wasteful in these scenarios.

- Concerning the number $K_{\mathrm{sam}}$ of interior sampling steps the results clearly indicate that using more interior sampling steps is more evaluation-efficient – but only for the *right* choice of interior sampler (NHR for the linear and modes problems; mRRT for the robotics problems). However, one should note that especially in the early phase one can see that this comes with less diversity-per-sample. E.g., the 6D modes problem profits from more frequent restarts (esp. with $\mathrm{dist}_{100}$ seeding).

- Finally, Fig. 8 confirms that injection of isometric noise during downhill is among the good methods, but the same method setting without injecting noise is better. Therefore, concerning downhill, in our evaluations aggressive GN slack steps without further noise or rejection (Wolfe or MH) mechanism performs well throughout the problems.

# 5 Conceptual Discussions

## 5.1 Is there a role for Lagrange Parameters in NLP Sampling?

For zero interior energy, $f(x) = 0$, any feasible point is a solution to the NLP with zero dual variable. As is obvious from the 1st KKT condition, the role of Lagrange parameters is to counteract cost gradients $-\nabla f(x)$ pulling into the infeasible region. Therefore, without costs, no Lagrange parameters. In other terms, if we only care about sampling feasible points (without energy), we can discard the whole idea of Lagrange parameters in the first place.

For non-zero costs, Lagrange parameters establish stationarity for a single optimum point located exactly on active constraints. This bears the question whether in NLP Sampling Lagrange parameters can equally play a role when sampling points, esp. points with active constraints. However, in view of our concrete proposed family of NLP Samplers, it seems the answer is negative. For instance, consider NHR sampling: The only difference between flat energy $f(x) = 0$ and non-zeros costs is the use of Metropolis-Hasting to calibrate the energies before and after a proposed step. At no point this method aims to sample exactly on active constraints. This argument seems to hold for any methods that truly samples in the *interior*, which leads to the question whether manifold sampling strategies could benefit from estimating Lagrange parameters. Such methods typically aim to move tangentially on the manifold and Lagrange parameters (e.g. added to the score of a Langevin method) could be used to steer gradient-based MCMC methods towards tangential steps. However, we can enforce tangential steps also much more explicitly using the constraint Jacobian (as mRRT does). Therefore, also for manifold samplers the potential role of estimated Lagrange parameters seems unclear. More fundamentally, also for manifold sampling we never care about a single point of stationarity, but rather about continuously making steps tangentially to the manifold.

Therefore, in view of the broad range of approaches considered in this paper, we currently do not see a role for Lagrange methods in NLP sampling.

## 5.2 Relaxed vs. Diffused NLP

We defined the *relaxed NLP* as $F_{\gamma\mu}(x) = \gamma f(x) + \mu s(x)^\top s(x)$, mixing square slack penalties with the energy. In contrast, our initial interested in this line of work was with the idea of a *diffused NLP:* Let $p(x_0) \propto e^{-f(x_0)}\ \mathbb{I}_{g(x_0)\le 0}\ \mathbb{I}_{h(x_0))=0}$ be our target distribution, where we renamed $x$ by $x_0$. A diffusion model describes the joint distribution $p(x_{0:T})$ as a Markov chain, where $p(x_0)$ is our target distribution, and each transition $p(x_{t+1}|x_t) = \mathcal{N}(x_t; \sqrt{\alpha_t} x_{t\text{-}1}, (1-\alpha_t)\mathbf{I})$ mixes Gaussian noise [10]. We have

$$p(x_t \,|\, x_0) = \mathcal{N}(x_t;\ \sqrt{\bar\alpha_t} x_0, (1-\bar\alpha_t)\mathbf{I})\ ,\quad \bar\alpha_t = \prod_{s=1}^t \alpha_t\ , \tag{5}$$

$$p(x_t) = \int p(x_t \,|\, x_0)\ p(x_0)\ dx_0\ . \tag{6}$$

We call $p(x_t)$ the *diffused NLP*. Based on this, there are two interesting questions: (1) Can we analytically describe or approximate $p(x_t)$? And (2), can we design an efficient NLP Sampler that follows the typical denoising process, starting to sample from $p(x_T)$ and denoising step-wise until we have a sample from $p(x_0)$?

Concerning (2), we failed to design an efficient NLP sampler following these lines of thoughts. The Appendix B sketches what we considered, but these approaches were orders of magnitude less efficient that the rather straight-forward two-phase methods proposed above. Especially, the aggressiveness and efficiency of GN steps in highly non-linear cases seemed to be lacking in our attempts.

Nevertheless, concerning (1), we can approximate the Diffused NLP $p(x_t)$ using local observations of $f, g, h$ and their gradients. Note that using linearizations of $f, g, h$ essentially means considering the local

Quadratic Program (assuming $f$ is a least squares energy); so what we describe here is a Diffused QP. In the appendix B we show that linear inequalities diffuse approximately to a product of sigmoidals

$$p_t(x) \approx \prod_i \Phi\Big[\frac{-\nabla g_i^\top x - c}{|\nabla g_i|\sigma}\Big] . \tag{7}$$

where $\Phi(x) = \int_{-\infty} \mathcal{N}(x, 0; 1)dx$ is the cumulative Gaussian transitioning from 1 (when $x$ is far inside the constraint) to 0 (when $x$ is far outside) and $\sigma, c$ are constants that depend on $t$ and the linearization of $g$. Note the gradient normalization: the Euclidean distance to the inequality hyperplane matters, not the value of $g_i(x)$. This approximation is highly accurate if constraints are linearly independent, see Fig. 9.

In contrast, the relaxed NLP (re-written as probability) includes factors

$$p_{\gamma\mu}(x_t) \propto \exp\{-F_{\gamma\mu}(x)\} \propto \prod_i \exp\{-\mu[g_i(x)]_+{}^2\} , \tag{8}$$

where $f$ and $h$ terms are neglected. This could be visualized as the right side of a Gaussian transitioning from 1 (when $g_i(x) = 0$) to 0 (when $g_i(x) \gg 0$). Therefore, the sigmoidal shape is slightly different to the Diffused NLP, but in particular also the scaling: Here the probability decreases with the scaled value of $g_i$, not the Euclidean distance to its hyperplane.

## 5.3 Globally Estimating the Partition Function

Consider a target distribution with two distant modes, e.g. with energy $F(x) = \min\{(x-x_1)^2/2+\alpha,\ (x-x_2)^2/2 + \beta\}$ in 1D, where the mode at $x_1$ has base-energy $\alpha$, and the mode at $x_2$ has base-energy $\beta$. In physics, these two modes could be macro states with different energy levels $\alpha$ and $\beta$. The energy levels determine how many samples (micro states) populate each mode. In our example (if $x_1$ and $x_2$ are distant) a fraction $w_1 = e^{-\alpha}\sqrt{2\pi}/Z$ of samples will be in the first mode, and $w_2 = e^{\beta}\sqrt{2\pi}/Z$ will be in the second mode.[5] The point here is that only observing a local energy value $F(x_1) = \alpha$ (or its gradient) provides no information on how much the mode is populated. Only the partition function $Z$ tells which *part* of the samples really populate a mode, namely $e^{-\alpha}\sqrt{2\pi}/Z$ in our case. In this view, discussing the partition function in sampling is analogous to the discussion of local vs. global optimization.

Our methods aim to efficiently discover modes, but do not correctly calibrate their weighting. That is, the "base of attraction" decides on the population of a mode, rather than its energy level relative to other modes. We want to note that within each mode, interior sampling methods ensure correct calibration of sample probabilities, but MCMC methods (with essentially zero mixing probability between modes) are no solution to calibrating across modes. If samples can be clearly clustered into modes, one approach to calibrating across modes could be to, in retrospect, re-weight samples based on their energies and density.

However, estimating the global partition function is beyond only calibrating across discovered modes: It implies estimating the total probability mass of modes we have not yet discovered at all. How to estimate this on the basis of our seeded restarting methods is an interesting challenge. It seems that global assumptions about the underlying functions (as made, e.g., in Bayesian optimization or other global optimization approaches) are necessary for an estimation, which is beyond the scope of this discussion. However, the above discussion helps to understand the gravity of globally correct sampling, which relates to our next discussion.

## 5.4 Diffusion Denoising Models vs. NLP Sampling

An initial motivation for this whole study was the impressive global coverage and diversity of samples generated by diffusion models such as DDPMs [10]. The generative process, Langevin dynamics guided by a model trained to predict the noise vector $\epsilon$ from the original sample $x_0$, has amazing properties in view of how it transports the seed density $p(x_T) = \mathcal{N}(x_T \mid 0, 1)$ to the target density $p(x_0)$. In particular, it correctly transports to all modes of the target density, with correct relative calibration. The intuitive picture behind the "Earth Mover Distance" works well here: Diffusion Denoising takes the seed density $p(x_T) = \mathcal{N}(x_T \mid 0, 1)$, splits its mass apart and moves it across the landscape, potentially uphill or across local dips of the target score of $p(x_0)$, so that eventually all probability mass is correctly moved to the final modes of the target density. How is this globally perfect transport of probability mass possible with DDPMs, when – in the view of the previous discussion – it seems excruciatingly hard within NLP Sampling?

[5] The $\sqrt{2\pi}$ comes from the Gaussian normalization, the partition function turns out $Z = \frac{\sqrt{2\pi}}{e^{-\alpha}+e^{-\beta}}$ and $w_1 = \sigma(\beta - \alpha)$, $w_2 = \sigma(\alpha - \beta)$.

The answer is data. DDPMs are trained on samples from the target density, which are (assumed) a correct global representation of all modes and their relative calibration. DDPMs do not have to discover modes; they are trained to reproduce, with exact same weighting, the modes in the data.

In contrast, in the NLP Sampling problem formulation we only have access to point-wise evaluations of $f, g, h$ with gradients – no global information is available at all.

In this view, NLP Sampling and learning generative models from data are very different. However, both are generative processes, and we established the close relations between the underlying methods. A core scientific question is if there is a possibility to design NLP samplers that realize a global transport similar to trained denoising models – a global transport from an initial seed density to the target density, covering all modes in a calibrated manner.

Our initial attempts to use the Diffused NLP to this end obviously (in retrospect) failed, as the Diffused NLP – when approximated from local evaluations of $f, g, h$ with gradients – contains no global information at all, and denoising on the Diffused NLP is as locally stuck as any other local MCMC or gradient based or SQP[6] method. Based on such local information, transport through local dips, global calibration, or even partition function estimation is out of reach. The family of methods studied here shows that – for a while – we gave up on the idea of realizing similar global sampling processes.

However, a promising avenue now seems to combine "data-free" sampling methods as proposed here with generative model learning. Training denoising models from samples and modes we have already discovered seems straight-forward, as demonstrated in existing work [21, 17]. An interesting question seems whether online trained denoising methods could also directly be leveraged and integrated within restarting two-phase samplers, rather than just alternating between them.

## 5.5 Sequential vs. Coordinated Sampling (e.g. Stein-Variational)

Our restarting two-phase methods generate samples sequentially: Each sample $x_i$ is generated by a run (involving downhill and interior sampling) that may be influenced by previous samples $D$. This is in contrast to parallel runs, where multiple particles descend in parallel to become samples of $p(x)$, as in Stein-Variational Gradient Descent [14].

On a conceptual level, it is interesting to compare the amount of information available for informed sample generation in a sequential vs. coordinated approach: In a sequential approach later runs are informed of actually feasible previous samples, whereas in parallel runs all particles inform each other but have no information yet about truly feasible regions. In both approaches, samples can "keep kernel distance", but in the parallel approach much more symmetrically.

While SVGD is guaranteed to eventually sample from $p(x)$, naively applying it on $p(x)$ (with zero value and score in the infeasible region) is prone to fail. However, there is a rich space of promising ideas to combine SVGD approaches with our framework: SVGD could be applied only for interior sampling and initialized with results from pure downhill runs. Or SVGD could be extended to include a downhill phase itself, where it is first applied on $F_{01}$ and then on $F_{1\mu}$. Finally, the combination of both ideas is interesting: batch-wise restarting of parallel runs, where particles are not only informed of each other but also the feasible samples found in the previous batch. This was also demonstrated in previous work on constrained SVGD for trajectory optimization [16], where it was demonstrated that taking the constraints into account while coordinating parallel runs using the SVGD update can be a powerful sampling method.

[6] Iteratively denoising the Diffused NLP is quite analogous to Sequential Quadratic Programming.

# 6 Summary

The aim of this study is to better understand approaches to efficiently generate diverse solutions to constrained problems. The fields of MCMC, constrained optimization, as well as robotics can all contribute ideas to this challenge. With this work we aimed to provide an integrative view and framework to combine methods from these fields, and gain insights in their strengths from the empirical evaluations.

On the technical side, our contributions are:

- We first proposed the NLP Sampling problem formulation, which is nothing but constrained sampling but under the typical assumptions of NLPs: the cost/energy and constraint functions can be accessed point-wise in a differentiable manner.
- We proposed an integrative framework, restarting two-phase NLP samplers, to combine methodologies flexibly.
- We proposed novel basic MCMC samplers, in particular Non-linear Metropolis-Adjusted Hit-and-Run (NHR), which extends HR to non-uniform interior energies using Metropolis-Hasting as well as to non-linear inequalities. Using a basic margin approach we can also this can also handle non-linear equalities when used in conjunction with slack reduction in each step.
- We proposed the Minimum Spanning Tree Score $\text{MSTS}_p$ as a novel metric to capture the diversity of generated samples. This metric greatly helped us to navigate through the large space of methods and conversely gain insights in the structure of our evaluation problems.

Beyond these technical contributions we hope that we also contributed on a conceptual level, starting with highlighting the strong relations, e.g., between Riemannian-Langevin and GN steps, or between Metropolis Hasting and the Armijo rule. Sec. 5 highlights that this study was particularly inspired by the impressive properties of data-trained denoising models, esp. its global transport of probability mass to correctly cover all modes. Our discussion mentions corresponding ideas, but also the negative results we found when trying to realize NLP Sampling by a Langevin process on a (locally approximated) Diffused NLP. Based on these insights we discussed how learning and diffusion models could help to tackle NLP sampling in the future.

# Appendix

## A Downhill, Langevin, Noise, and Step Rejection

In this section we consider an energy function $F(x)$. As explained in Sec. 3.1 this may be the pure slack downhill energy $F_{01}$ or the interior sampling energy $F_{1\mu}, \mu \gg 1$.

**Downhill Steps & Noise:** The first ingredient of downhill methods is a *plain gradient step* $-\alpha\nabla F(x)$ or a *Gauss-Newton (GN) step* $-\alpha(\nabla^2 F(x) + \lambda\mathbf{I})^{-1}\nabla F(x)$ with GN-approximated Hessian $\nabla^2 F(x)$ and regularization $\lambda\mathbf{I}$. We generally post-process these steps by clipping with a potential maximal stepsize $\delta_{\text{max}}$ and accounting for box constraints.

Noise injection is uncommon in classical optimization, but an interesting candidate to increase robustness to small local optima, plateaus or only piece-wise continuous gradients (as common in robotics applications). *Isotropic noise* adds Gaussian noise,

$$x \leftarrow x + \sigma z \ , \quad z \sim \mathcal{N}(0, \mathbf{I}_n) \ , \tag{9}$$

with standard deviation $\sigma \in \mathbb{R}$ as parameter. *Covariant noise* adds Gaussian noise with precision matrix equal to the regularized Hessian $H = \nabla^2 F_{\gamma\mu}(x) + \lambda\mathbf{I}$,

$$x \leftarrow x + \sigma\sqrt{H^{-1}} z \ , \quad z \sim \mathcal{N}(0, \mathbf{I}_n) \ , \tag{10}$$

where $\sqrt{H^{-1}}$ refers to the Cholesky decomposition of $H^{-1}$.

The use of $H^{-1}$ as noise covariance is motivated by interpreting the Hessian as a local metric relative to which the Gauss-Newton step is the covariant gradient. Using covariant noise for slack minimization $F_{0\mu}$ has the following effects: (1) Directions without slack have small eigenvalue $\kappa + \lambda$ and therefore large standard deviation $\sqrt{1/(\kappa+\lambda)}$. That is, the noise explores more along directions without slack, and is smaller in directions with slack. (2) In the limit $\mu \to \infty$, $\sqrt{H^{-1}}$ has zero eigenvalues along active directions. In this case, the noise become strictly tangential to slack isolines. Therefore, covariant noise realizes many aspects we would intuitively see as beneficial for manifold sampling and exploration along slack isolines.

**Relation to Langevin variants:** Given a distribution $p(x) = \frac{1}{Z} e^{-F(x)}$ with energy $F(x)$ and partition function $Z$, the score function is neg-energy gradient

$$\mathbf{s}(x) = \nabla_x \log p(x) = \nabla_x \log \frac{1}{Z} e^{-F(x)} = -\nabla_x F(x) \ , \tag{11}$$

which is independent of $Z$. *Langevin dynamics* in continuous time is described by

$$\dot{x} = -\nabla F(x) + \sqrt{2}\xi \tag{12}$$

with Brownian motion $\xi$. For a discrete time (Euler integration) step, this translates to

$$x_{t+1} = x_t - \tau\nabla F(x_t) + \sqrt{2\tau} z \ , \quad z \sim \mathcal{N}(0, 1) \tag{13}$$

where the variance of the Brownian noise step only increases with $\sqrt{\tau}$. In this view, what is special about Langevin is the *relative calibration* of stepsize and noise. We summarize this as:

**Observation 1.** *Discrete time Langevin dynamics (13) combines plain gradient steps $-\alpha\delta$ with isotropic noise where the noise scaling $\sigma = \sqrt{2\alpha}$ is tied to the step size $\alpha \equiv \tau$.*

*Riemann Manifold Langevin* (Girolami et al, 2011) is described by the continuous time dynamics

$$\dot{x} = -G(x)^{-1}\nabla F(x) + \sqrt{2G(x)^{-1}}\xi \ , \tag{14}$$

where $G(x)$ is a non-Euclidean metric in $x$-space. The work of [8] considers $x$ to be a parameter of another probability distribution $p(y;x)$ (e.g., when MCMC is used to sample in hierarchical Bayes models), in which case $G(x)$ is naturally the Fisher Information metric, which is the Hessian of the relative entropy, $G(x_0) = \nabla^2{}_x D\big(p(y;x) \,\|\, p(y;x_0)\big)$. But in general, $G(x)$ can be any non-Euclidean metric. Note that if defining the *steepest descent direction* in a non-Euclidean case as

$$\delta^* = \operatorname*{argmin}_\delta \nabla f(x)^\top \delta \quad \text{s.t.} \quad \delta^\top G \delta = 1 \ , \tag{15}$$

then the (covariant) solution is $\delta^* = -G^{-1}\nabla f(x)$. And if $x$ are distribution parameters and $G$ chosen to be the Fisher information metric, this is called natural gradient descent direction. Both are equivalent to a Newton step, identifying the Hessian $\nabla^2 f(x)$ with a non-Euclidean metric. We summarize this as:

**Observation 2.** *Discrete time Riemannian Langevin dynamics combines Newton downhill steps $-\alpha H^{-1}\delta$ with covariant noise where the noise scaling $\sigma = \sqrt{2\alpha}$ is still tied to the step size $\alpha \equiv \tau$.*

**Rejecting Steps with Armijo & Metropolis-Hasting:** Both, the field of classical optimization and the field of MCMC, offer a core paradigm for step rejection: Backtracking line search in the first case, and Metropolis-Hasting in the second. While both are utterly important and central in their respective fields, I am not aware of discussions of their relation.

The 1st Wolfe condition in backtracking line search is to accept a step from $x$ to $x' = x + \alpha\delta$ if

$$F(x') \le F(x) + \rho \nabla F(x)^\top (x' - x) \ , \tag{16}$$

with parameter $\rho$ typically $\in [0.01, 0.1]$ [2]. Note that for a plain gradient step $\delta = -\nabla F(x)$ this reduces to $F(x') \le F(x) - \rho\alpha\delta^2$, which is called *Armijo rule.* These rules describe a "sufficient" decrease of the objective. (We omit discussing a second typical acceptance criterion, called 2nd Wolfe or curvature condition, describing sufficient decrease of the directional derivative.)

In contrast, in *Metropolis-Hasting (MH)* we accept a step from $x$ to $x'$ with probability

$$\min\{1, \frac{p(x')}{p(x)} \, \frac{q(x|x')}{q(x'|x)}\} \ , \tag{17}$$

where $q(x'|x)$ is the prior proposal step probability, and $p(x) \propto \exp\{-F(x)\}$. The second term relates to an asymmetry of a-priori likelihood of steps from $x$ to $x'$ versus from $x'$ to $x$. Typically, we would assume the forward step probability $q(x'|x)$ to be moderately high for any $x'$ we indeed sampled. However, the inverse step probability $q(x|x')$ may easily become very small if we have little noise but directional bias in the proposal.

Consider a Langevin-type step $x' = x - \alpha\delta + \sigma z$ which combines a downhill step $\delta$ with isometric noise $z$ (with stepsize $\alpha$ and noise sdv $\sigma$). For a 2nd step, away from $x'$, we use the notation $x'' = x' - \alpha\delta' + \sigma z'$. Then

$$q(x'|x) \propto \exp\{-\tfrac{1}{2\sigma^2}(x' - x + \alpha\delta)^2\} \tag{18}$$

$$q(x|x') \propto \exp\{-\tfrac{1}{2\sigma^2}(x - x' + \alpha\delta')^2\} \ . \tag{19}$$

We find that the most likely step $x' = x - \alpha\delta$ has a MH ratio

$$\frac{q(x|x')}{q(x'|x)} = \exp\{-\tfrac{1}{2\sigma^2}(\alpha(\delta + \delta'))^2\} \ \le 1 \tag{20}$$

$$\frac{p(x')}{p(x)} \, \frac{q(x|x')}{q(x'|x)} = \exp\{-[F(x') - F(x) + \tfrac{2\alpha^2}{\sigma^2}\bar\delta^2]\} \ , \quad \bar\delta = \tfrac{\delta+\delta'}{2} \ . \tag{21}$$

This MH ratio is $\ge 1$ (which guarantees accept) if

$$F(x') \le F(x) - \tfrac{2\alpha^2}{\sigma^2}\bar\delta^2 \ . \tag{22}$$

This relates strongly to the Armijo rule $F(x') \le F(x) - \rho\alpha\delta^2$ for the plain gradient step $\delta = \nabla F(x)$ and $\rho = \frac{2\alpha}{\sigma^2}$. We summarize this as:

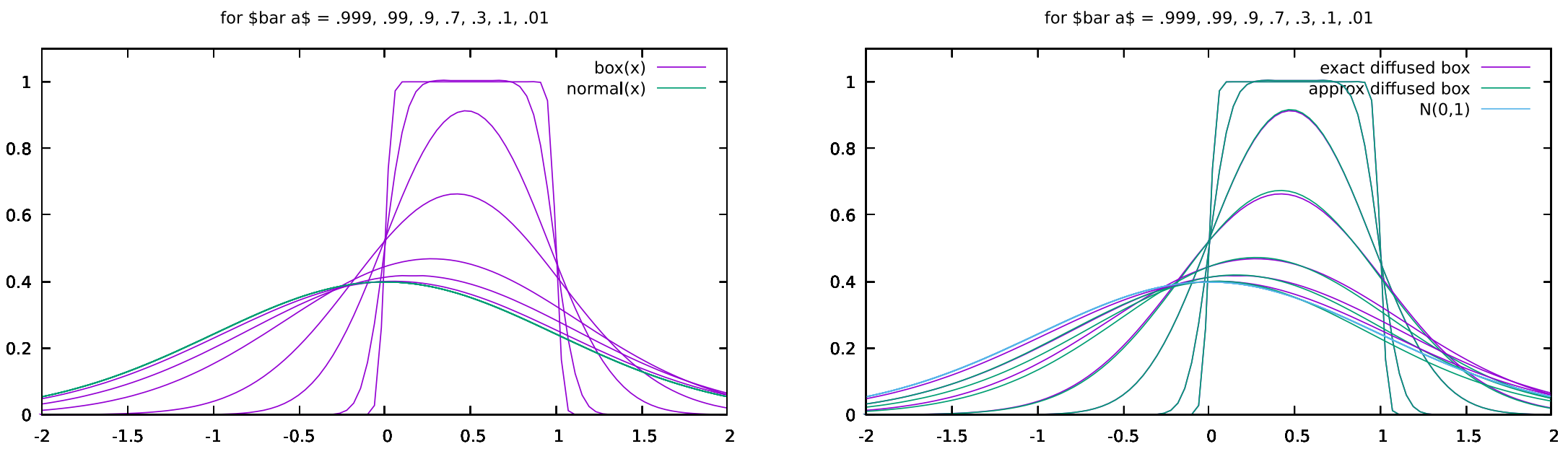

Figure 9

**Observation 3.** *The acceptance rate ([21]) of a gradient downhill step* $x' = x - \alpha\delta$ *under Metropolis Hasting differs from the Armijo rule with line search parameter* $\rho = \frac{2\alpha}{\sigma^2}$ *in that it uses the more symmetrical gradient estimate* $\bar\delta$ *instead of* $\delta$*, and accepts with some probability* $< 1$ *also steps with* $F(x') > F(x) - \frac{2\alpha^2}{\sigma^2}\delta^2$*.*

Note that when scaling $F(x) \leftarrow \mu F(x)$ for $\mu \to \infty$, also the MH rule would lead to strict acceptance based on the inequality ([22]).

# B Relation between Relaxed NLP and Diffused NLP

Sec. 5.2 discusses using the Diffused NLP instead of the Relaxed NLP as a basis for NLP Sampling. The Diffused NLP is the distribution $p(x_t)$ that arises from taking our $p(x) \propto \exp(-f(x))\ \mathbb{I}_{g(x)\le 0}\ \mathbb{I}_{h(x)=0}$ and diffusing it following DDPMs [10]. The following derives an approximation of this $p(x_t)$.

**1D Interval case:** Let's first investigate a 1D interval case, where $p(x_0), x_0 \in \mathbb{R}$ is defined via the NLP

$$f(x) = 0 \ \text{ s.t. } \ l \le x \le u \ . \tag{23}$$

As $t$ is fixed, we introduce simplifying notation $a = \sqrt{\bar\alpha_t}$, $\sigma^2 = (1-\bar\alpha_t)\mathbf{I}$, so that ([5]) becomes $p(x_t \,|\, x_0) = \mathcal{N}(x_t; a x_0, \sigma^2)$, and the convolution is

$$\begin{aligned} p(x_t) &= \int p(x_x \,|\, x_0)\ p(x_0)\ dx_0 = \frac{1}{u-l}\int_l^u p(x_x \,|\, x_0)\ dx_0 && (24)\\ &= \frac{1}{u-l}\int_l^u \mathcal{N}(x_t; a x_0, \sigma^2)\ dx_0 = \frac{1}{a(u-l)}\int_{al}^{au} \mathcal{N}(x_t; y, \sigma^2)\ dy && (25)\\ &= \frac{1}{a(u-l)}\Big[\Phi(\frac{au - x_t}{\sigma}) - \Phi(\frac{al - x_t}{\sigma})\Big]\ . && (26) \end{aligned}$$

where $\Phi(x) = \int_{-\infty} \mathcal{N}(x, 0; 1) dx$ is the cumulative Gaussian. Figure 9 plots this for various $\bar\alpha$, converging to the standard Gaussian for $\bar\alpha \to 0$.

Note that this is a *difference* of two cumulative Gaussians. But generalization to multiple higher-dimensional inequalities is much simpler if this was a *product* of sigmoid-shaped functions, where each constraint contributes a factor to $p(x_t)$. We note that this difference of cumulative Gaussians can well be approximated as a product,

$$\Phi(y) - \Phi(z) \approx \Phi(y)\Phi(-z)\ . \tag{27}$$

Fig. 9 (right) displays the diffusion when the exact difference is replaced by this product.

**General inequalities:** We can generalize this approximation to arbitrary inequalities in $\mathbb{R}^n$. The $l, u$-bounds above can be rewritten as inequalities $g_1(x) = l - x$ and $g_2(x) = x - u$, respectively. Therefore, they contribute factors of the form $\Phi(z)$ with $z = \frac{x-al}{\sigma} = \frac{-ag_1(x)+(1-a)x}{\sigma}]$. Consider

$$f(x) = 0 \ \text{ s.t. } \ g(x) \le 0 \ , \tag{28}$$

where $p(x_0)$ is uniform in the feasible polytope. Using a linear approximation $g_i(x) \approx \nabla g_i(x')^\top (x - x') + g_i(x')$ each inequality contributes a factor

$$p_t(x) \propto \prod_i \Phi\Big[\frac{-ag_i(x) - (1-a)\nabla g_i(x)^\top x}{|\nabla g_i(x)|\sigma}\Big] \approx \prod_i \Phi\Big[\frac{-\nabla g_i(x')^\top (x - ax') - ag_i(x')}{|\nabla g_i(x')|\sigma}\Big] \ . \tag{29}$$

Note that for $a = 1$ the argument is the (gradient-normalized) original inequality, while for $a = 0$ the offset of the hyperplane is shifted to go through zero – the parameter $a$ therefore interpolates between the sigmoidal function going through the original inequality offset and going through zero.

Within application of hit-and-run, we will consider $p(x_t)$ along a random direction $d$, where $x = \hat{x} + \beta d$. In this case, we have

$$p_t(x) \approx \prod_i \Phi\Big[\frac{-\nabla g_i(x')^\top (\hat{x} + \beta d - ax') - ag_i(x')}{|\nabla g_i(x')|\sigma}\Big] \tag{30}$$

$$= \prod_i \Phi\Big[\frac{\beta - b_i}{s_i}\Big] \quad \text{with} \quad \frac{1}{s_i} = \frac{-\nabla g_i(x')^\top d}{|\nabla g_i(x')|\sigma} \ , \quad b_i = s_i \frac{\nabla g_i(x')^\top (\hat{x} - ax') + ag_i(x')}{|\nabla g_i(x')|\sigma} \tag{31}$$

**General Equality:** Consider

$$f(x) = 0 \ \text{ s.t. } \ h(x) \le 0 \ , \tag{32}$$

where $p(x_0)$ is uniform along the feasible manifold $h(x) = 0$. Eq. (5) described the exact diffusion of a point $x_0$, namely to $x_t = ax_0 + z$ with $z \sim \mathcal{N}(0, \sigma^2)$, where the mean is interpolated from $x_0$ to zero with decreasing $a$. For the equality case, we consider a local linearization $h(x) = h(x') + J_h(x')(x - x')$ with Jacobian $J_h$, and describe the diffusion of the feasible hyperplane by decomposing along tangential directions (along which the distribution would be uniform), and normal directions.

Let $P_h = J_h^\top (J_h J_h^\top)^1 J_h$ be the projection into the subspace normal to the hyperplane. The diffusion in this normal direction is generated by $x_t = ax_0 + P_h z$ with $z \sim \mathcal{N}(0, \sigma^2)$. Given the local linearization around $x'$, we choose $x_0$ as the nearest projection $x_0 = x' - J_h^\top (J_h J_h^\top)^1 h(x')$